\documentclass[10pt,twocolumn,letterpaper]{article}

\usepackage{cvpr}              
\usepackage{float}
\usepackage{placeins}
\definecolor{cvprblue}{rgb}{0.21,0.49,0.74}
\usepackage[pagebackref,breaklinks,colorlinks,allcolors=cvprblue]{hyperref}
\usepackage{tabularx}
\usepackage{multirow}
\def\confName{CVPR}
\def\confYear{2026}

\title{Roots Beneath the Cut: Uncovering the Risk of Concept Revival in Pruning-Based Unlearning for Diffusion Models}

\usepackage{amssymb} 
\usepackage{amsmath}

\author{
    \begin{tabular}{c}
        Ci Zhang$^{1\dagger}$ \quad Zhaojun Ding$^{1\dagger}$ \quad Chence Yang$^{1}$ \quad Jun Liu$^{2,3}$ \quad Xiaoming Zhai$^{1}$ \\
        Shaoyi Huang$^{4}$ \quad Beiwen Li$^{1}$ \quad Xiaolong Ma$^{5}$ \quad Jin Lu$^{1}$ \quad Geng Yuan$^{1}$ \\[0.5em]
        $^{1}$University of Georgia \quad $^{2}$Carnegie Mellon University \\
        $^{3}$Northeastern University \quad
        $^{4}$Stevens Institute of Technology \\ $^{5}$University of Arizona \\
        {\small \{zhangci, geng.yuan\}@uga.edu}
    \end{tabular}
    \thanks{$^\dagger$Equal contribution.}
}

\begin{document}
\maketitle

\newcommand{\GY}[1]{\textcolor{blue}{Geng: #1}}
\newcommand{\changed}[1]{\textcolor{blue}{#1}}
\newcommand{\bred}[1]{\textcolor{red}{\sf\bfseries #1}}
\newcommand{\tred}[1]{\textcolor{red}{#1}}
\newcommand{\blue}[1]{\textcolor{blue}{#1}}
\newcommand{\yellow}[1]{\textcolor{yellow}{#1}}
\newcommand{\purple}[1]{\textcolor{purple}{#1}}
\newcommand{\brown}[1]{#1}
\newcommand{\cross}[1]{\textcolor{red}{\sout{#1}}}

\begin{abstract}
Pruning-based unlearning has recently emerged as a fast, training-free, and data-independent approach to remove undesired concepts from diffusion models. It promises high efficiency and robustness, offering an attractive alternative to traditional fine-tuning or editing-based unlearning. However, in this paper we uncover a hidden danger behind this promising paradigm. We find that the locations of pruned weights, typically set to zero during unlearning, can act as side-channel signals that leak critical information about the erased concepts.
To verify this vulnerability, we design a novel attack framework capable of reviving erased concepts from pruned diffusion models in a fully data-free and training-free manner. Our experiments confirm that pruning-based unlearning is not inherently secure, as erased concepts can be effectively revived without any additional data or retraining. Extensive experiments on diffusion-based unlearning based on concept related weights lead to the conclusion: once the critical concept-related weights in diffusion models are identified, our method can effectively recover the original concept regardless of how the weights are manipulated.  
Finally, we explore potential defense strategies and advocate safer pruning mechanisms that conceal pruning locations while preserving unlearning effectiveness, providing practical insights for designing more secure pruning-based unlearning frameworks. Code is available at \url{https://github.com/Brankozz/Roots-Beneath-the-Cut}.

\end{abstract}    
\section{Introduction}

Recent advances in text-to-image diffusion models have achieved remarkable success in generating high-quality, semantically aligned images, driving their widespread adoption across research and creative applications~\cite{rombach2022high,ma2024sit,podell2023sdxl,chen2023pixart,deepfloydif}.
However, these models are typically trained on massive datasets~\cite{schuhmann2022laion,schuhmann2021laion,changpinyo2021conceptual} that inevitably contain sensitive, private, or copyrighted information, raising serious privacy, ethical, and legal concerns~\cite{schramowski2023safe, kim2023towards, li2025responsible}. Regulations such as the European Union’s General Data Protection Regulation (GDPR)~\cite{voigt2017gdpr} and the California Consumer Privacy Act (CCPA)~\cite{delatorre2018ccpa} grant individuals the “right to be forgotten,” requiring the removal of personal data from trained models.

To address these challenges, machine unlearning has emerged as a promising solution to selectively erase unwanted information without full retraining, ensuring privacy compliance while preserving model capability~\cite{cao2015towards, ginart2019making, gandikota2023erasing, gandikota2024unified, kim2023towards, kumari2023ablating, bourtoule2021machine, liu2025rethinking, kurmanji2023towards}. 
It aims to efficiently ensure privacy compliance while preserving the model’s general capabilities. Unlearning methods for diffusion models include editing-based approaches~\cite{kumari2023ablating, gandikota2023erasing, gandikota2024unified, zhang2024forget, orgad2023editing, lu2024mace}, which modify latent or token representations to suppress specific concepts, and training-based approaches~\cite{heng2023selective, zhao2024separable, liu2024implicit, wu2024erasediff, fan2023salun, li2025sculpting}, which fine-tune the model using tailored loss functions to remove undesired information. While these methods have demonstrated effectiveness, they are often computationally expensive and difficult to scale.

In contrast, \textbf{pruning-based unlearning}~\cite{jia2023model,chavhan2024conceptprune,pochinkov2024dissecting,li2025sculpting,romandini2025fedup,xiao2025right} has recently emerged as a highly promising paradigm that offers a fundamentally different perspective. By identifying and removing the weights associated with specific concepts, pruning-based methods achieve concept erasure in a completely training-free manner. 
They offer high efficiency, robustness to adversarial prompts, and minimal degradation in generative quality, making pruning-based unlearning an increasingly practical and scalable solution for large diffusion models~\cite{li2025sculpting}.

\begin{figure}{}
    \label{fig:intro}
    \centering
    \includegraphics[width=0.8 \linewidth]{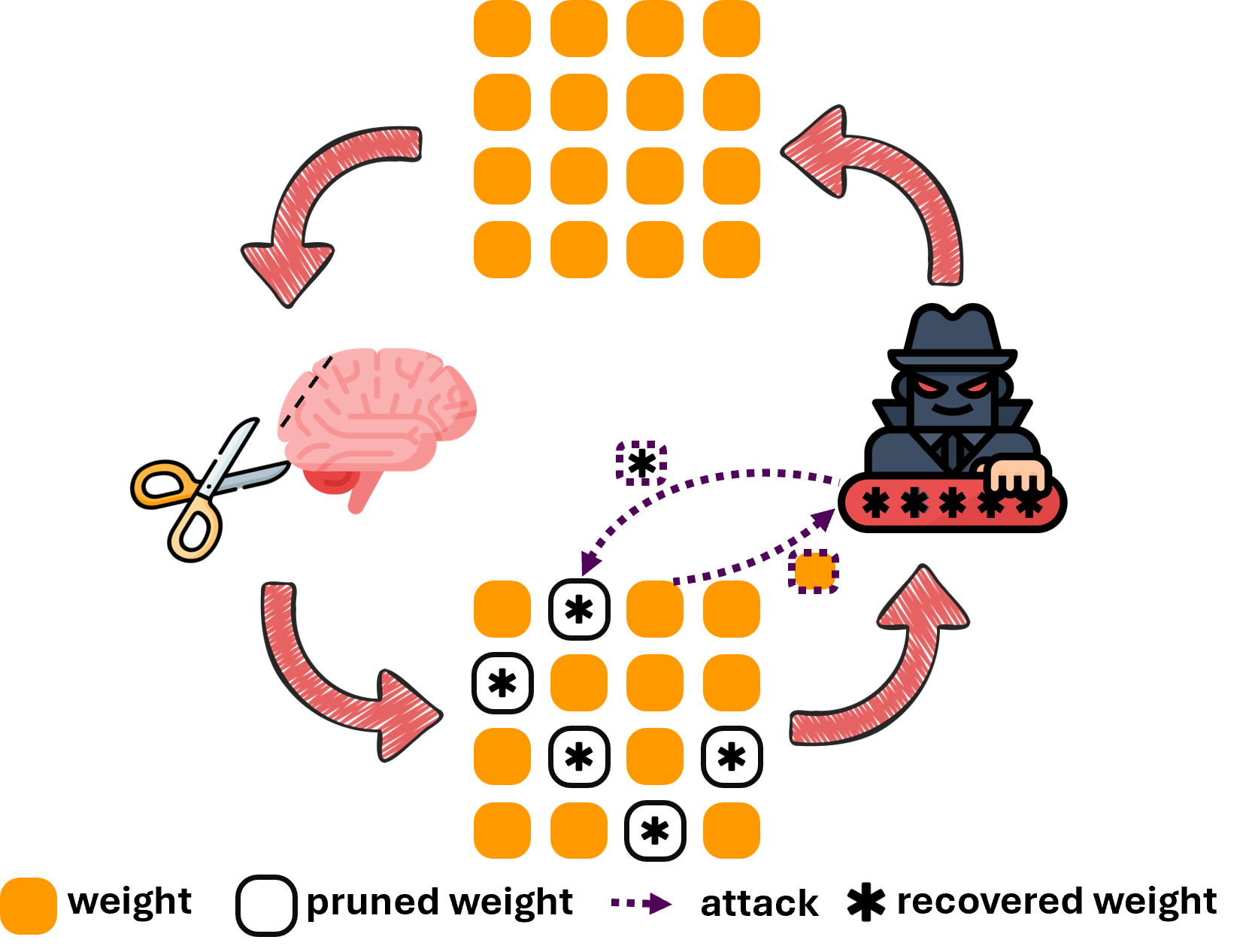}
    \vspace{-2mm}
    \caption{Inherent risk of pruning-based machine unlearning.}
    \label{fig:attack_sparsity}
    \vspace{-3mm}
\end{figure}

However, while people have embraced the effectiveness and efficiency of pruning-based unlearning methods, a critical security concern has been largely overlooked.
As hinted by the title \textit{Roots Beneath the Cut}, pruning may not completely remove the underlying knowledge, leaving residual traces that can be recovered.
These methods typically erase concepts by setting the associated weights to zero, but such visible pruning locations reveal where key concept-related parameters once existed. Consequently, this information can act as side-channel information exploitable by potential attackers (as shown in Fig.~\ref{fig:attack_sparsity}).
It is therefore essential to investigate pruning-based unlearning for its inherent vulnerabilities and newly exposed attack surface. This leads to a crucial question:
\textit{\textbf{Can an attacker, under a data-free and training-free setting, and without access to the original magnitudes of the pruned weights, exploit pruning location information alone to reconstruct these weights and revive the supposedly erased concepts?}}

To address these questions, we begin by examining the relative importance of recovering the signs versus the magnitudes of the pruned weights. Our analysis reveals a key insight: restoring the weight signs alone, even without precise magnitude recovery, can substantially revive the erased concepts.
Building upon this observation, we design an attack framework that approximates the recovery of pruned weights in the unlearned diffusion model. 
Specifically, the framework is designed first to incorporate a low-rank matrix completion module to estimate the original signs of the pruned weights, followed by a Top-K Sign Retention module that only preserves high-confidence signs. Finally, we propose a Neuron-Max Scaling strategy to assign appropriate magnitudes to the recovered weights.

Our experimental results show that the proposed framework can successfully recover more than 70\% of the signs of the pruned weights, which is sufficient to revive the erased concepts. Compared with unlearned models, our attack restores the accuracy of erased concepts from an average of 8\% to 54\% within only seven minutes, without requiring any data or retraining.
We validate our framework across multiple unlearning tasks, including object unlearning, artistic-style unlearning, and NSFW-content unlearning. The results consistently demonstrate its effectiveness in all scenarios, significantly reviving erased concepts while maintaining the model’s generative quality.
These findings confirm that although pruning-based unlearning offers strong efficiency and practicality, it inevitably introduces a critical attack surface that can be effectively exploited, revealing a previously overlooked security vulnerability. This work underscores the necessity of reexamining the security assumptions underlying pruning-based unlearning and developing safer frameworks for future diffusion models.

In addition, our work advocates that future pruning-based unlearning methods avoid directly setting pruned weights to zero. Instead, we propose a simple yet effective defense that replaces zeroed weights with Gaussian noise of controlled variance, effectively obscuring pruning locations and preventing exploitation. However, if the variance is too small, the pruning positions may still be identifiable, while excessive variance can degrade unlearning performance. To balance this trade-off, we analyze how the variance level influences unlearning effectiveness and provide a case study that offers insights and guidance for developing more secure pruning-based unlearning methods. We hope this simple defense serves as an initial step toward inspiring future work on improving the security of pruning-based unlearning.


\noindent\textbf{Contributions of this paper are summarized as follows.}
\begin{itemize}
\item We are the first to identify a previously overlooked security risk in pruning-based unlearning, showing that the locations of pruned weights may serve as side-channel information that could potentially be exploited to recover erased visual concepts.

\item We develop a novel data-free and training-free attack framework that confirms this vulnerability can indeed be exploited in practice, successfully reviving erased concepts from pruned diffusion models.

\item Extensive experiments across object, artistic-style, and NSFW-content unlearning tasks demonstrate the effectiveness of our attack framework, which recovers over 70\% of pruned-weight signs and improves the average accuracy of erased concepts from 8\% to 54\%.

\item We explore potential defense strategies and advocate for safer pruning mechanisms by introducing a bounded-Gaussian pruning defense that conceals pruning locations while preserving unlearning effectiveness, providing practical insights for designing more secure pruning-based unlearning frameworks.

\end{itemize}

\section{Related Work}
\noindent\textbf{Machine Unlearning for Diffusion Models}. 
Machine unlearning was first introduced to make trained models behave as if they had never seen certain data such as private or copyrighted content \cite{7163042}. Subsequent studies \cite{bourtoule2021machine, graves2021amnesiac, guo2019certified} refined its theoretical foundations and formal definitions. With the rapid progress of generative AI, unlearning has gained growing attention in text-to-image diffusion models, where it enables selective removal of undesired concepts while preserving the overall generative quality.

Among existing approaches, some methods achieve unlearning by directly modifying latent or token representations to suppress target concepts \cite{kumari2023ablating, gandikota2023erasing, zhang2024forget, orgad2023editing, lu2024mace}. These editing-based techniques often rely on similarity maximization or token remapping to align the representation of an erased concept with that of a retained one. However, their performance is highly sensitive to the choice of reference concept, which affects both forgetting quality and preservation of benign concepts \cite{bui2025fantastic}. Selecting an optimal mapping target typically requires exploring the entire concept space, leading to substantial computational overhead.
Other approaches remove concepts through fine-tuning with customized loss functions that encourage the model to forget the target concept while retaining general generative capabilities \cite{heng2023selective, zhao2024separable, liu2024implicit, wu2024erasediff, fan2023salun, li2025sculpting}. Although these training-based methods can achieve strong erasure performance, they are computationally expensive and often require retraining on large datasets. Moreover, recent works have revealed that adversarial textual prompts can still circumvent several editing-based techniques once considered robust \cite{kumari2023ablating, gandikota2023erasing, zhang2024forget}, exposing their potential vulnerability.

To overcome these limitations, pruning-based unlearning methods \cite{jia2023model,chavhan2024conceptprune,pochinkov2024dissecting,li2025sculpting,romandini2025fedup,xiao2025right} have emerged as training-free and data-independent alternatives for efficient concept removal. Specifically for diffusion models, they remove neurons or weights linked to target concepts, achieving concept erasure with minimal impact on generation quality. ConceptPrune \cite{chavhan2024conceptprune} effectively removes object categories, art styles, gender, and nudity while maintaining fidelity and diversity. These methods also show robustness to adversarial prompts and offer an efficient, scalable paradigm for secure diffusion model deployment. Sculpting Memory \cite{li2025sculpting} further extends this idea with dynamic masking and concept-aware optimization for stable multi-concept forgetting.

\section{Motivations and Preliminaries}
\label{Sec:Motivations}

Pruning-based unlearning removes weights associated with specific concepts, but it remains unclear which aspects of the pruned weights, such as their locations, signs, or magnitudes, retain recoverable information that may enable concept restoration. To examine whether these residual traces can be exploited, we conduct a preliminary investigation on diffusion models pruned by ConceptPrune~\cite{chavhan2024conceptprune}, focusing on how the signs and magnitudes of the pruned weights respectively affect the recovery of erased visual concepts.

\begin{figure}{}
    \centering
    \includegraphics[width=0.8\linewidth]{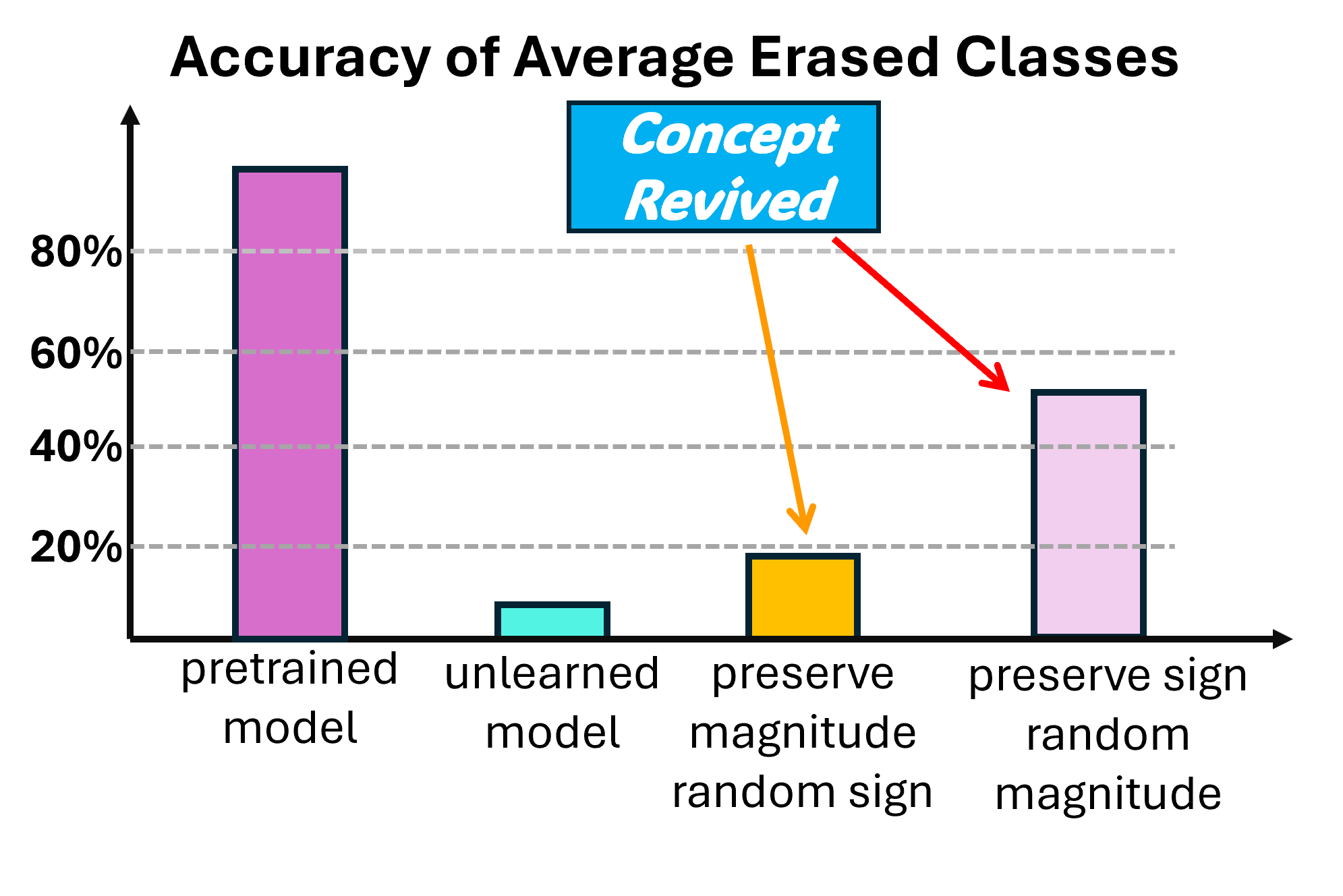}
    \vspace{-2mm}
    \caption{Restored accuracy on erased concept class.}
    \label{fig:thrust2_preliminary}
    \vspace{-3mm}
\end{figure}

We perform a study under an idealized setting where either the signs or the magnitudes of the pruned weights are assumed to be perfectly recoverable. Specifically, we compare two scenarios: precise magnitudes with random signs, and precise signs with random magnitudes. The results, as shown in Fig.~\ref{fig:thrust2_preliminary}, reveal a key finding: the correctness of weight signs plays a far more critical role in reviving erased concepts than the accuracy of their magnitudes.

We further evaluate different magnitude assignment strategies when the signs are correctly recovered. By assigning mean or max magnitudes from the corresponding remaining neurons, we find that the erased concepts can indeed be partially recovered under certain conditions. These results indicate that the pruning process may leave sufficient residual information for an adversary to exploit and reconstruct supposedly erased concepts. In particular, accurately restoring the signs proves substantially more important for concept revival than recovering precise magnitudes (more details in Appendix A). 
This finding naturally leads to the next question: \textit{\textbf{how can we accurately infer and reconstruct both the signs and magnitudes of pruned weights in a training-free setting?}}

\begin{figure*}[t]
    \centering
    \includegraphics[width=0.85\textwidth]{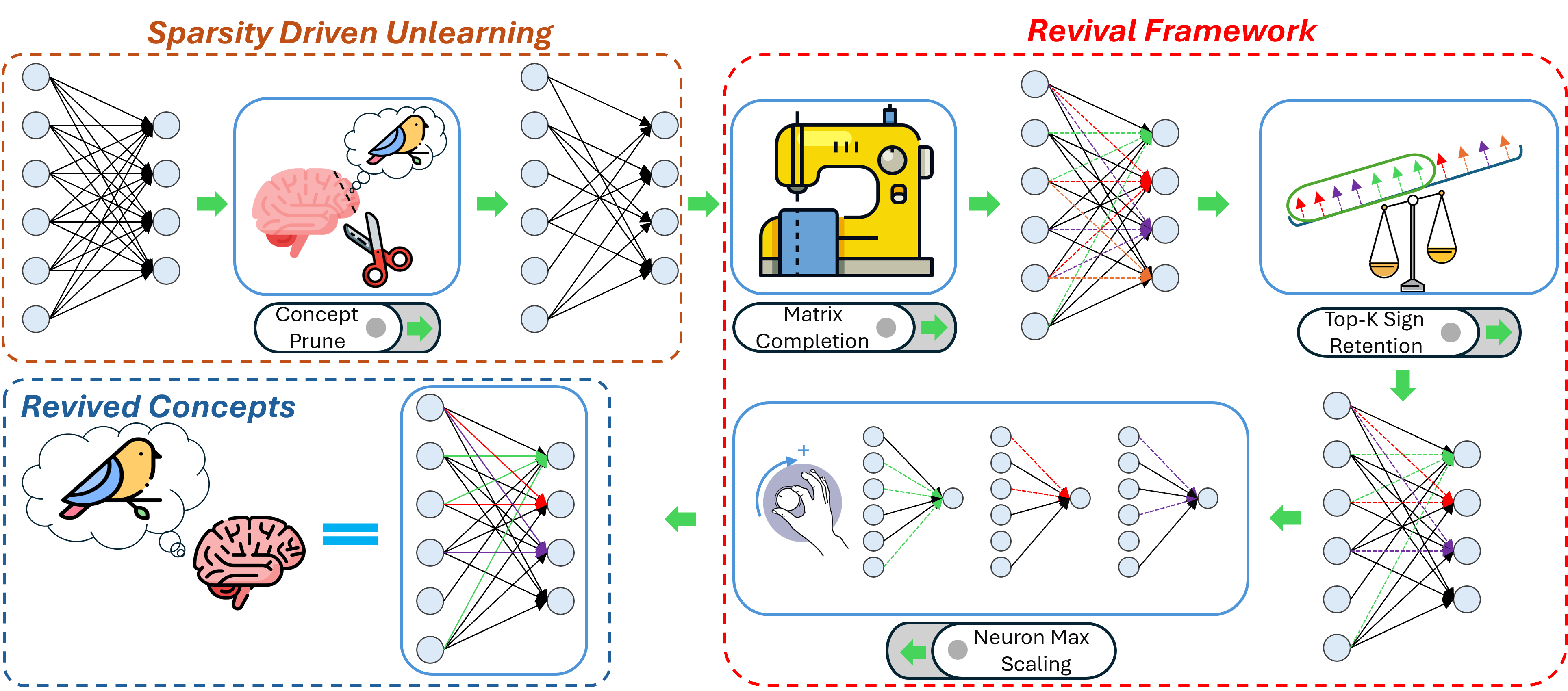}
    \vspace{-2mm}
    \caption{Overview of sparsity driven unlearning and our proposed revival framework.}
    \vspace{-4mm}
    \label{fig:revival_framework}
\end{figure*}

\section{Methodology}
Building upon the insights found in Section~\ref{Sec:Motivations}, we next introduce our attack framework that leverages the remaining network parameters to infer missing signs and values, enabling effective training-free and data-free revival of erased visual concepts.
The red area in Fig. \ref{fig:revival_framework} demonstrates the structure of our framework, which is based on the Matrix Completion, Top-K Sign Retention, and Neuron Max Scaling detailed in sections \ref{matrix_completion}, \ref{topk_retention}, and \ref{neuron_max_scaling}, respectively. Besides, we propose a defense strategy in Sec. \ref{sec:defense_method}.

\subsection{Low-rank Matrix Completion}
\label{matrix_completion}

Motivated by prior work in low-rank matrix recovery~\cite{cai2010singular,mazumder2010spectral}, which demonstrates that under appropriate low-rank assumptions, one can infer the values of unobserved matrix entries by solving a nuclear norm regularized reconstruction problem. Although such methods cannot perfectly recover the exact magnitudes of missing weights in our setting, we empirically observe that they provide \emph{surprisingly strong recovery of weight signs}, which is precisely the information most critical for reviving erased concepts. Consequently, the first stage of our revival framework applies matrix completion to obtain an approximate reconstruction of the pruned weights, from which sign information is subsequently extracted and refined in Sec.~\ref{topk_retention}.

Inspired by nuclear-norm regularized matrix completion method Iterative Soft-Thresholded SVD (IST-SVD)~\cite{cai2010singular}, considering an incomplete matrix $X \in \mathbb{R}^{m \times n}$, with observed entries indexed by $\Omega$. Matrix completion solves:
\begin{equation}
    \min_{M} \frac{1}{2}\|P_\Omega(X) - P_\Omega(M)\|_F^2 + \lambda \|M\|_*,
    \label{eq:mc_objective}
\end{equation}
where $P_\Omega(\cdot)$ projects onto observed entries, and $\|\cdot\|_*$ denotes the nuclear norm, a convex surrogate for matrix rank.

Then, we iteratively reconstructs missing entries using the current estimate, followed by a singular value decomposition (SVD) and a soft-thresholding operation applied to the singular values:
\begin{equation}
    \mathcal{S}_\lambda(Y) = 
    U \, \mathrm{diag}((\sigma_i - \lambda)_+) \, V^\top,
    \label{eq:soft_thresholding}
\end{equation}
where $Y = U \Sigma V^\top$ is the SVD and $(t)_+ = \max(t,0)$ is the soft-thresholding operator.  
This step enforces a low-rank structure by shrinking small singular values toward zero. 
Although conceptually simple and theoretically grounded, it becomes prohibitively expensive for large diffusion models since each iteration may require a full SVD of dense matrices containing millions of parameters.

To overcome this issue, we incorporate \textit{SoftImpute} algorithm~\citep{mazumder2010spectral} was proposed as a scalable and more efficient variant tailored for large-scale matrix completion. It solves the same convex objective in Eq.~\ref{eq:mc_objective}, but introduces a more structured and memory-efficient iteration scheme.

At iteration $t$, we replaces the missing entries of $X$ by the corresponding entries of the current low-rank estimate $M^{(t)}$, yielding:
\begin{equation}
    Z^{(t)} = P_\Omega(X) + P_{\Omega^c}(M^{(t)}),
    \label{eq:softimpute_fill}
\end{equation}
where $\Omega^c$ indexes missing entries.  
Then, it performs SVD on $Z^{(t)}$ and applies soft-thresholding.

SoftImpute introduces two major practical advantages:
\begin{enumerate}
    \item \textbf{Sparse+Low-Rank structure.}  
    The matrix $Z^{(t)}$ is never materialized as a dense matrix; instead, its observed and imputed components are handled separately. This allows efficient use of truncated SVD solvers, significantly reducing both memory footprint and runtime.
    \item \textbf{Regularization path with warm starts.}  
    Solutions for larger $\lambda$ values serve as initializations for smaller ones, substantially accelerating convergence. As shown in~\citep{mazumder2010spectral}, SoftImpute enjoys global convergence with an $O(1/k)$ rate.
\end{enumerate}

We further implement SoftImpute in Python with GPU-accelerated truncated SVD, enabling efficient reconstruction of parameter matrices at diffusion-model scale.
Despite being unable to precisely reconstruct weight magnitudes, SoftImpute reliably recovers a substantial portion of their \emph{signs}. As shown in Sec.~\ref{topk_retention}, these recovered signs form the foundation of our Top-K Sign Retention strategy. Thus, matrix completion provides an effective first stage of our revival pipeline and a strong initial estimate for restoring erased concepts.

\subsection{Top-K Sign Retention}
\label{topk_retention}

Since matrix completion cannot perfectly recover all weight signs, we seek to mitigate the impact of incorrect ones. We first examine whether sign correctness correlates with magnitude and observe that large-magnitude weights in the recovered results tend to align with those in the pretrained model. Moreover, weights with higher magnitudes are more likely to have correct signs, as detailed in Appendix~B. These observations suggest that our method can effectively recover the most important subset of missing weights.

Building on this insight, we introduce the magnitude-based Top-K Sign Retention module (Fig.~\ref{fig:revival_framework}), which preserves the signs of the top-K recovered weights with the largest magnitudes while setting the rest to zero. The retained weights are then assigned the maximum magnitude within their corresponding neurons. Zeroing out small-magnitude weights reduces outlier effects and significantly improves recovery performance, as shown later in Table~\ref{tab:ten_class}. This approach intuitively emphasizes high-confidence signs, preventing large errors from incorrectly recovered weights and enhancing overall concept revival quality.

\subsection{Neuron-Max Scaling}
\label{neuron_max_scaling}
Based on the insights observed in the ideal case experiment, when all signs are correctly recovered, assigning the maximum magnitude from the remaining weight neuron can further enhance concept revival compared to other magnitude assignment strategies. This naturally motivates us to adopt the same strategy on top of the \textbf{Top-K Sign Retention} procedure. Nevertheless, we also tried alternative magnitude assignment strategies, including the mean magnitude within each remaining weight neuron, sampled values from remaining weight neurons' distribution, and remaining layer weights' distribution. 

Among these strategies, we consistently find that assigning the maximum magnitude achieves the best revival performance across different Top-K on sign retention, which is demonstrated in Sec. \ref{sec:ablation}. Based on this observation, we formalize this procedure as a neuron-level magnitude scaling method, which we term \textbf{Neuron-Max Scaling (NMS)} shown in the red frame of Fig. \ref{fig:revival_framework}. This component amplifies the performance brought by signs retained to the important subset of the pruned weights by Top-K Sign Retention, effectively restoring the most influential activation pattern during the inference.




\subsection{Gaussian Obfuscation Defense for Pruning-Based Unlearning}
\label{sec:defense_method}

We advocate that future pruning-based unlearning methods should conceal the identifiable pruning footprint to prevent exposure of pruned weight locations. To this end, we propose a simple yet effective defense mechanism that hides these traces while preserving the unlearning effect.


Our key idea is to replace the pruned weights with values sampled from a carefully chosen smooth distribution that mimics the natural statistics of the unmodified parameters. Empirically, we observe that the weight distribution of each layer is approximately symmetric around zero, with a unimodal shape and rapidly decaying tails, which can be well approximated by a zero-centered Gaussian.
Thus, to hide the pruning footprints, we draw each pruned weight from a Gaussian distribution 
\(\mathcal{N}(0,\sigma_M^2)\), 
rather than assigning zero. 
This ``Gaussian obfuscation'' maintains the visual indistinguishability of modified and unmodified weights, preventing mask detection and significantly improving resistance to sign-based recovery attacks.

To analyze when the Gaussian-obfuscated pruned weights are statistically indistinguishable from 
unmodified entries, we consider the following setup. 
Let the unmodified parameter values follow density \(f_U(x)\), and the modified (i.e., Gaussian-filled) 
values follow density \(f_M(x)\). 
Let \(\alpha\in(0,1)\) denote the global fraction of modified entries. 
Given an interval \(I=[\ell,u]\), the probability that a randomly selected value lying in \(I\) is modified is
\begin{align}
  p_I
  &= \mathbb{P}(\text{modified}\mid X\in I) \notag\\
  &= \frac{\alpha\int_{\ell}^{u} f_M(x)\,dx}{
        \alpha\int_{\ell}^{u} f_M(x)\,dx
        + (1-\alpha)\int_{\ell}^{u} f_U(x)\,dx } .
  \label{eq:prob-general-defense}
\end{align}

\noindent
\textbf{Symmetric-Normal Specialization.}
We now adopt the empirical observation that both unmodified and modified values are well-approximated 
by centered normal distributions:
\begin{align}
  X_U &\sim \mathcal{N}(0,\sigma_U^2), \notag\\
  X_M &\sim \mathcal{N}(0,\sigma_M^2), \notag\\
  I &= [-w,w],\quad w>0. 
  \label{eq:normal-assumptions-defense}
\end{align}
The probability mass of a centered Gaussian within interval \([-w,w]\) is
\begin{align}
  \int_{-w}^{w}\frac{1}{\sqrt{2\pi}\sigma}
    e^{-\frac{x^2}{2\sigma^2}}\,dx
  &= 2\Phi\!\Big(\frac{w}{\sigma}\Big)-1
  = \operatorname{erf}\!\Big(\frac{w}{\sqrt{2}\sigma}\Big).
  \label{eq:normal-mass-defense}
\end{align}
Substituting Eq.\eqref{eq:normal-mass-defense} into Eq.\eqref{eq:prob-general-defense} yields
\begin{align}
  p(w)
  &= \frac{\alpha\,\operatorname{erf}(\tfrac{w}{\sqrt{2}\sigma_M})}{
        \alpha\,\operatorname{erf}(\tfrac{w}{\sqrt{2}\sigma_M})
        + (1-\alpha)\,\operatorname{erf}(\tfrac{w}{\sqrt{2}\sigma_U}) } .
  \label{eq:prob-normal-defense}
\end{align}

The term \(p(w)\) quantifies the likelihood that values in an interval centered around zero originated 
from the obfuscation process. Importantly, Eq.\eqref{eq:prob-normal-defense} reveals a fundamental 
\textit{security--utility tradeoff}:
When \(\sigma_M\) is extremely small, the modified values cluster tightly near zero, making pruning 
highly effective, but simultaneously making the modified values statistically easier to detect. 
Conversely, a large \(\sigma_M\) hides the pruned values more effectively but eventually disrupts the 
model's generative capability.  
This tradeoff motivates the experiments in Sec.~\ref{sec:defense_experiment}, where we study the effect 
of \(\sigma_M\) on both pruning performance and mask detectability.

Finally, we emphasize that our defense mechanism is lightweight and fully compatible with existing 
pruning-based unlearning pipelines. In practice, we simply replace each zeroed weight with a sample from 
\(\mathcal{N}(0,\sigma_M^2)\), with no retraining or re-optimization required.  
Theoretical guidance from Eq.\eqref{eq:prob-normal-defense}, combined with empirical validation, enables the practitioner to select \(\sigma_M\) that balances unlearning fidelity and resistance to recovery attacks.

\section{Experiments}

\begin{table*}[tp]
    \centering
    \caption{Top-1 classification accuracy of erased and preserved objects, using a pre-trained ResNet-50. All neuron methods are based on Top-0.6 Sign Retention.}
    \vspace{-1mm}
    \resizebox{\textwidth}{!}{
    \begin{tabular}{lccccccc|cccccc}
        \toprule
        \multirow{2.5}{*}{\textbf{Classes} } 
        & \multicolumn{6}{c}{\textbf{Accuracy of Erased Class}} 
        &
        & \multicolumn{6}{c}{\textbf{Accuracy of Preserved Class} $\uparrow$} 
        \\
        \cmidrule(lr){2-7} \cmidrule(lr){9-14}

        & \multirow{2.5}{*}{\textbf{\shortstack{Pretrained~\cite{rombach2022high}\\SD-v1.5}}} 
        & \multirow{2.5}{*}{\textbf{\shortstack{Concept~\cite{chavhan2024conceptprune}\\Prune $\downarrow$}}} 
        & \multirow{2.5}{*}{\textbf{\shortstack{Quant~\cite{zhang2024catastrophic}\\Recover $\uparrow$}}} 
        & \multirow{2.5}{*}{\textbf{\shortstack{Neuron\\Sample $\uparrow$}}}
        & \multirow{2.5}{*}{\textbf{\shortstack{Neuron\\Average $\uparrow$}}}
        & \multirow{2.5}{*}{\textbf{\shortstack{NMS\\Ours $\uparrow$}}}
        &
        & \multirow{2.5}{*}{\textbf{\shortstack{Pretrained~\cite{rombach2022high}\\SD-v1.5}}} 
        & \multirow{2.5}{*}{\textbf{\shortstack{Concept~\cite{chavhan2024conceptprune}\\Prune}}} 
        & \multirow{2.5}{*}{\textbf{\shortstack{Quant~\cite{zhang2024catastrophic}\\Recover}}}
        & \multirow{2.5}{*}{\textbf{\shortstack{Neuron\\Sample}}}
        & \multirow{2.5}{*}{\textbf{\shortstack{Neuron\\Average}}}
        & \multirow{2.5}{*}{\textbf{\shortstack{NMS\\Ours}}}
        
        \\
        &
        & & & & 
        \\
        \midrule

        Church & 0.86 & 0.08 & 0.07 & 0.17 & 0.18 & \textbf{0.63}  && 0.94 & 0.78 & 0.79 & 0.80 & 0.81 & \textbf{0.91} \\
        Golf Ball & 0.81 & 0.16 & 0.19 & 0.26 & 0.25 & \textbf{0.56}  && 0.93 & 0.88 & 0.82 & 0.86 & 0.87 & \textbf{0.89} \\
        Gas Pump & 0.828 & 0.07 & 0.08 & 0.14 & 0.14 & \textbf{0.31} && 0.94 & 0.81 & 0.81 & 0.86 & 0.88 & \textbf{0.90} \\
        Parachute & 0.94 & 0.06 & 0.08 & 0.11 & 0.13 & \textbf{0.71}  && 0.93 & 0.74 & 0.75 & 0.87 & 0.89 & \textbf{0.91} \\
        Chain Saw & 0.71 & 0.00 & 0.02 & 0.01 & 0.00 & \textbf{0.40}  && 0.95 & 0.82 & 0.83 & 0.82 & 0.83 & \textbf{0.95} \\
        French Horn & 0.99 & 0.02 & \textbf{0.36} & 0.02 & 0.04 & 0.26  && 0.92 & 0.81 & 0.81 & 0.86 & 0.84 & \textbf{0.92} \\
        Mountain Bike & 0.97 & 0.01 & 0.00 & 0.03 & 0.03 & \textbf{0.58}  && 0.93 & 0.75 & 0.76 & 0.76 & 0.77 & \textbf{0.89} \\
        Starfish & 1.00 & 0.32 & 0.31 & 0.49 & 0.52 & \textbf{0.80}  && 0.93 & 0.79 & 0.79 & 0.88 & 0.89 & \textbf{0.90} \\
        Spider Web & 0.96 & 0.22 & 0.21 & 0.46 & 0.36 & \textbf{0.69}  && 0.93 & 0.83 & 0.83 & 0.83 & 0.83 & \textbf{0.91} \\
        School Bus & 0.99 & 0.02 & 0.00 & 0.07 & 0.06 & \textbf{0.60}  && 0.93 & 0.76 & 0.73 & 0.74 & 0.77 & \textbf{0.88} \\
        Racket & 0.97 & 0.04 & 0.03 & 0.08 & 0.07 & \textbf{0.45}  && 0.93 & 0.83 & 0.84 & 0.86 & 0.86 & \textbf{0.92} \\
        Candle & 0.97 & 0.01 & 0.03 & 0.13 & 0.14 & \textbf{0.48}  && 0.93 & 0.77 & 0.77 & 0.83 & 0.84 & \textbf{0.92} \\
        \midrule
        \textbf{Average} 
        & 0.927 & 0.08 & 0.12 & 0.16 & 0.16 & \textbf{0.54} 
        &&
        0.93 & 0.80 & 0.79 & 0.83 & 0.84 & \textbf{0.91} \\
        \bottomrule
    \end{tabular}}
    \vspace{-3mm}
    \label{tab:ten_class}
\end{table*}

\subsection{Experiment settings}
\label{experimentsettings}
\noindent\textbf{Experiment Details:} We strictly follow the experimental setup of ConceptPrune~\cite{chavhan2024conceptprune} to ensure a fair and consistent comparison. Specifically, we use the \textbf{Stable Diffusion v1.5} model, with 16 FFN layers served as candidates for pruning. For all subsequent unlearning tasks, we adhere to the same settings as ConceptPrune to obtain the pruned models. We then apply our proposed recovery framework to each FFN layer of the pruned model individually, recovering the complete model, which is subsequently used for concept revival and evaluation across different unlearning tasks.

\noindent\textbf{Baseline:} Among existing attack methods on machine unlearning, \textbf{Quant Recover}~\cite{zhang2024catastrophic} is the \emph{only} non-finetuning and training-free approach. Although it was originally proposed for large language models, its underlying principle of using low-bit quantization to obscure minimal weight changes during unlearning can be naturally extended to diffusion models. For a fair comparison, we apply the same 4-bit quantization protocol from~\cite{zhang2024catastrophic} to the same pruned models obtained from  ConceptPrune~\cite{chavhan2024conceptprune} and adopt the resulting quantized models as a baseline for evaluating our revival framework.

\vspace{-1mm}
\subsection{Revive Erased Object}
\label{objects}
\begin{figure*}
    \centering
    \includegraphics[width=0.9\textwidth]{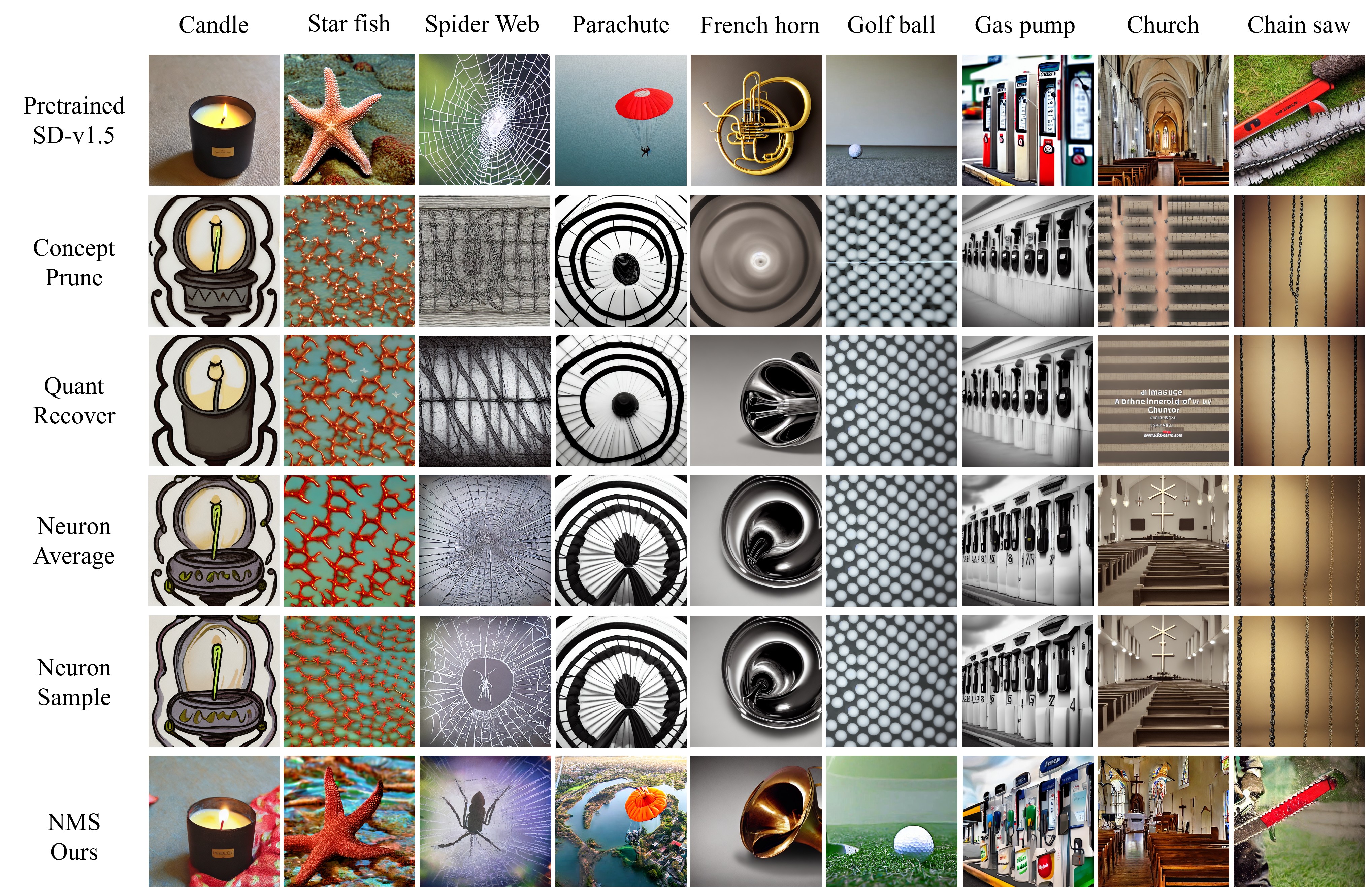}
    \caption{Visual results for twelve erased object classes. We compare the pretrained Stable Diffusion model, the unlearned model by ConceptPrune~\cite{chavhan2024conceptprune}, and four revival methods. Our NMS restores high-fidelity object appearances and preserves scene coherence, outperforming Quant Recover~\cite{zhang2024catastrophic}, Neuron Average, and Neuron Sample.}
    \label{fig:visual_results}
\end{figure*}

We conduct our evaluation experiments on the \textbf{ImageNet} dataset \cite{5206848}. To more comprehensively demonstrate the capability of our framework in concept revival, we adopt 12 classes in a subset from ImageNet. For each class, we generate 500 images via the same settings of ConceptPrune~\cite{chavhan2024conceptprune} and evaluate the revival performance with a pretrained \textbf{ResNet-50} classifier by measuring the top-1 classification accuracy.
In addition, we report the accuracy of each class across other five models: the pretrained \textbf{SD-v1.5} model, the standard unlearned model by ConceptPrune~\cite{chavhan2024conceptprune}, the Quant Recover~\cite{zhang2024catastrophic} baseline, Neuron Average and Neuron Sample, in which neuron average and neuron sample are both derived from the signs restored by our NMS procedure. Neuron Average assigns values using the mean magnitude of the remaining weights connected to each neuron, whereas Neuron Sample draws values by sampling from the empirical distribution of the remaining magnitudes. As shown in Table~\ref{tab:ten_class}, our method effectively revives the unlearned object concepts while preserving the model’s ability to generate other unrelated concepts. More results in Appendix E demonstrate that our NMS attack can recover original concepts once the concept related weights are identified regardless of how they are manipulated. The visualization results are shown in Fig \ref{fig:visual_results}.


\begin{figure}[b]
    \centering
    \vspace{-2mm}
    \hspace*{-8mm} 
    \includegraphics[width=\linewidth]{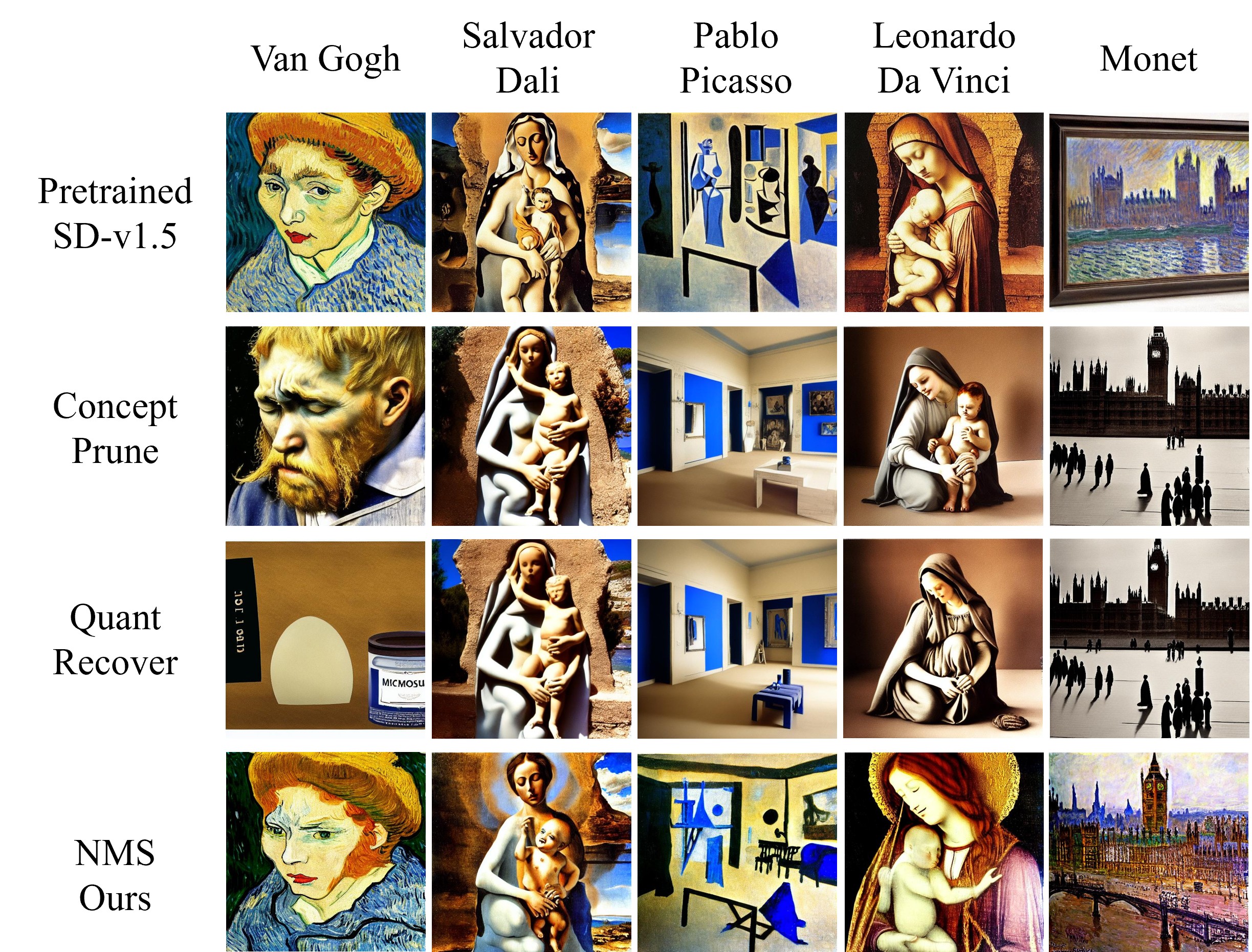}
    \caption{Visual results of revived art styles.}
    \label{fig:revive_artstyle}
    \vspace{-4mm}
\end{figure}

\subsection{Revive Erased Art Style}
\label{artstyle}

Since artistic styles often involve copyright concerns in real-world applications, we select five widely recognized artists (\textbf{Van Gogh}, \textbf{Monet}, \textbf{Pablo Picasso}, \textbf{Leonardo da Vinci}, and \textbf{Salvador Dali}) for evaluation. Following the experimental setup of ConceptPrune~\cite{chavhan2024conceptprune}, we use 50 prompts for each artist generated by ChatGPT, such as “\textit{Almond Blossoms by Vincent van Gogh}.”
We adopt the same evaluation metrics: \textbf{CLIP similarity}, which measures semantic alignment between the generated image and prompt, and \textbf{CLIP score}, which penalizes cases where the unlearned model yields higher similarity than the pretrained SD model. Lower CLIP similarity and higher CLIP score indicate stronger unlearning, while effective revival shows the opposite trend.
To assess overall generation fidelity, we also compute the \textbf{FID} and \textbf{CLIP similarity} on the COCO30K dataset~\cite{lin2014microsoft}.
We evaluate four methods: pretrained SD-v1.5, the unlearned model by ConceptPrune~\cite{chavhan2024conceptprune}, the Quant Recover~\cite{zhang2024catastrophic}, and the model recovered by our Neuron-Max Scaling (NMS).
Table~\ref{tab:artist_coco_comparison} shows that our framework \textbf{successfully revives} the erased artistic styles, achieving higher alignment with the original concepts while preserving overall generative quality. The corresponding visual results are presented in Fig.~\ref{fig:revive_artstyle}.

\begin{table*}[t]
\centering
\caption{Comparison between SD-V1.5, Concept Prune~\cite{chavhan2024conceptprune}, Quant Recover~\cite{zhang2024catastrophic} and NMS (Ours) on artist style and Coco datasets.}
\label{tab:artist_coco_comparison}
\resizebox{\textwidth}{!}{%
\begin{tabular}{l|c|c|c|c|cccc|cccc|cccc}
\toprule
\multirow{3}{*}{\textbf{Artist}} 
& \multicolumn{4}{c}{\textbf{SD-v1.5~\cite{rombach2022high}}} 
& \multicolumn{4}{c}{\textbf{Concept Prune~\cite{chavhan2024conceptprune}}}
& \multicolumn{4}{c}{\textbf{Quant Recover~\cite{zhang2024catastrophic}}}
& \multicolumn{4}{c}{\textbf{NMS (Ours)}} \\
\cmidrule(lr){2-5} \cmidrule(lr){6-9} \cmidrule(lr){10-13} \cmidrule(lr){14-17}
& \multicolumn{2}{c}{\textbf{Artist style}} & \multicolumn{2}{c}{\textbf{COCO~\cite{lin2014microsoft}}}
& \multicolumn{2}{c}{\textbf{Artist style}} & \multicolumn{2}{c}{\textbf{COCO~\cite{lin2014microsoft}}}
& \multicolumn{2}{c}{\textbf{Artist style}} & \multicolumn{2}{c}{\textbf{COCO~\cite{lin2014microsoft}}}
& \multicolumn{2}{c}{\textbf{Artist style}} & \multicolumn{2}{c}{\textbf{COCO~\cite{lin2014microsoft}}} \\
\cmidrule(lr){2-3} \cmidrule(lr){4-5}
\cmidrule(lr){6-7} \cmidrule(lr){8-9}
\cmidrule(lr){10-11} \cmidrule(lr){12-13}
\cmidrule(lr){14-15} \cmidrule(lr){16-17}
& Similarity & Score & FID & Similarity 
& Similarity $\downarrow$ & Score $\uparrow$ & FID $\downarrow$ & Similarity $\uparrow$
& Similarity $\uparrow$ & Score $\downarrow$ & FID $\downarrow$ & Similarity $\uparrow$
& Similarity $\uparrow$ & Score $\downarrow$ & FID $\downarrow$ & Similarity $\uparrow$ \\
\midrule
Van Gogh & 0.33 & \multirow{5}{*}{0} &\multirow{5}{*}{18.4} & \multirow{5}{*}{0.31} & 0.27 & 0.90 & 20.80 & 0.32 & 0.27 & 0.88 & 22.62 & 0.31 & \textbf{0.33} & \textbf{0.38} & \textbf{18.57} & 0.31  \\
Picasso  & 0.29 &  &  &  & 0.23 & 0.66 & 21.00 & 0.26 & 0.26 & 0.74 & 19.74 & 0.31 & \textbf{0.27} & \textbf{0.58} & \textbf{19.52} & 0.31  \\
Monet & 0.34 &  &  &  & 0.22 & 1.0 & 25.42 & 0.30 & 0.22 & 1.0 & 25.47 & 0.30 & \textbf{0.31} & \textbf{0.76} & \textbf{18.97} & \textbf{0.31}  \\
Da Vinci & 0.29 &  &  &  & 0.25 & 0.82 & 20.61 & 0.31 & 0.26 & 0.74 & 20.22 & 0.31 & \textbf{0.27} & \textbf{0.56} & \textbf{18.98} & 0.31  \\
Dali & 0.31 &  &  &  & 0.26 & 0.7 & 20.24 & 0.31 & 0.30 & 0.78 & 23.74 & 0.30 & \textbf{0.33} & \textbf{0.26} & \textbf{18.63} & 0.31  \\
\midrule
\textbf{Average} 
& 0.31 & 0 & 18.4 & 0.31
& 0.25 & 0.82 & 21.45 & 0.30
& 0.26 & 0.83 & 22.36 & 0.31
& \textbf{0.30} & \textbf{0.51} & \textbf{18.93} & 0.31 \\
\bottomrule
\end{tabular}}
\end{table*}

\subsection{Revive Explicit Content}
\label{nsfw}
We quantitatively evaluate our proposed revival framework on restoring Not-Safe-for-Work (NSFW) concepts that have been erased from the model. We adopt three benchmark prompt sets: the Inappropriate Prompts Dataset (I2P, 4,703 prompts) \cite{schramowski2023safe}, MMA (1,000 prompts) \cite{yang2024mma}, and Ring-A-Bell (101 prompts) \cite{tsai2023ring}. Using the nudity detector in \cite{bedapudi2025nudenet}, the pretrained SD-v1.5 model is found to trigger 461, 57, and 22 nudity detections on I2P, MMA, and Ring-A-Bell, respectively. After specific unlearning using ConceptPrune \cite{chavhan2024conceptprune}, these numbers drop substantially to 74, 57, and 22. However, our revival framework significantly revived the erased NSFW concepts, raising the detection counts to 118, 172, and 57, respectively. These results clearly demonstrate the high efficacy of our framework in reviving NSFW-related concepts that were previously erased. More results are in Appendix C.

\subsection{Defense Evaluation via Gaussian Obfuscation}
\label{sec:defense_experiment}

We now empirically evaluate the effectiveness of our Gaussian obfuscation defense. 
Following Sec.~\ref{sec:defense_method}, we replace the zeroed parameters in the pruning mask 
with values sampled from a zero-mean Gaussian distribution \(\mathcal{N}(0,\sigma_M^2)\), 
and systematically vary \(\sigma_M\) to study its impact on (i) pruning performance and 
(ii) detectability of modified entries.



\begin{figure}[t]
    \centering
    \includegraphics[width=0.9\linewidth]{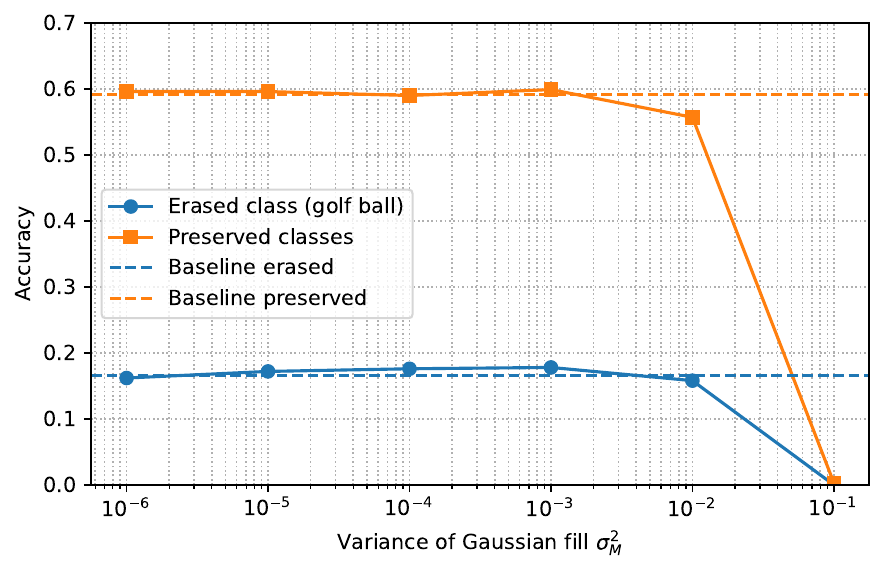}
    \vspace{-2mm}
    \caption{Effect of Gaussian obfuscation variance on unlearning performance.}
    \label{fig:defense_variance_curve}
    \vspace{-6mm}
\end{figure}

\noindent
\textbf{Effect on Pruning Performance.}
We evaluate our defense using the erased concept \textit{golf ball}.
After applying the standard pruning-based unlearning procedure, the removed weights are filled with Gaussian noise of varying variances $\sigma_M^2 \in \{10^{-6}, 10^{-5}, \dots, 10^{-1}\}$.
For each variance, we measure accuracy on both the erased and preserved classes.
As shown in Fig.~\ref{fig:defense_variance_curve}, both accuracies remain close to the pruning baseline for small variances, indicating that replacing zeros with weak Gaussian noise does not compromise unlearning.
When the variance becomes large, the samples deviate from the natural parameter distribution, degrading overall performance.
These results reveal a practical regime where pruning traces can be effectively hidden while maintaining unlearning fidelity.
This behavior aligns with the analytical tradeoff in Eq.~\eqref{eq:prob-normal-defense}.
Smaller \(\sigma_M\) preserves pruning quality but increases detectability, while larger \(\sigma_M\) enhances concealment at the cost of generation quality. Our experiments identify the region where this tradeoff is effectively balanced.

\begin{figure}[t]
    \centering
    \includegraphics[width=0.8\linewidth]{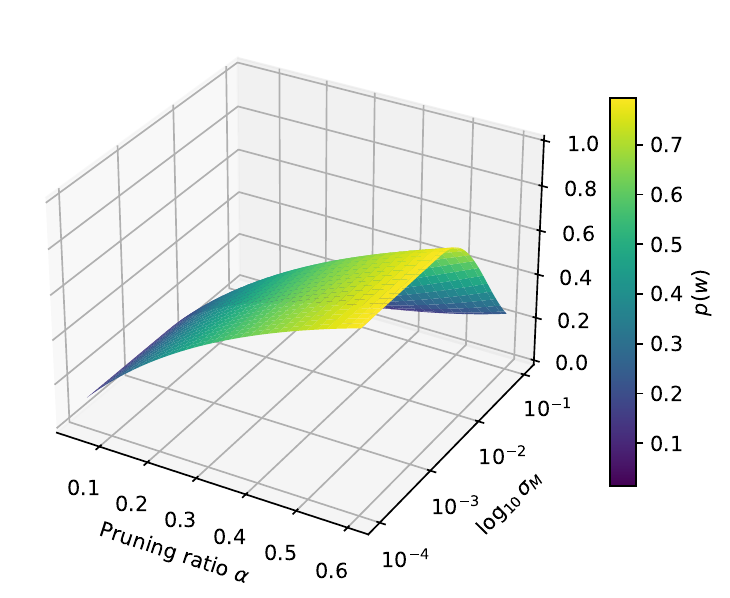}
    \caption{
    The conditional probability \(p(w)\) from Eq.~\eqref{eq:prob-normal-defense}
    as a function of pruning ratio \(\alpha\) and obfuscation variance \(\sigma_M\),
    with \(\sigma_U\) fixed.  
    This surface provides practical guidance for selecting \(\sigma_M\) to balance 
    concealment and unlearning performance.}
    \label{fig:defense_surface}
\end{figure}

\noindent
\textbf{Guidance for Choosing \texorpdfstring{$\sigma_M$}{sigmaM}.}
To further illustrate how the analytical expression in Eq.~\eqref{eq:prob-normal-defense} informs defense design, we visualize the conditional probability \(p(w)\) as a function of 
\(\alpha\) (the pruning ratio) and \(\sigma_M\), while fixing \(\sigma_U\) according to empirical measurements of unmodified weights.
Figure~\ref{fig:defense_surface} depicts the resulting surface.  
The plot reveals regions of high detectability (large \(p(w)\)) and 
low detectability (small \(p(w)\)).  
While \(\sigma_U\) is fixed in this visualization, practitioners may estimate it 
from model statistics, and then consult the theoretical landscape to select a 
\(\sigma_M\) that achieves the desired concealment level for a given pruning ratio~\(\alpha\).

\label{sec:experiments}

\vspace{-2mm}
\subsection{Ablation Study}
\label{sec:ablation}

We conduct an ablation study to clearly illustrate how different combinations of \textit{Top-k Sign Retention} and magnitude assignment strategies affect the revival performance. In this analysis, we use the \textbf{golf ball} unlearning task as an example for concept revival. As shown in Table~\ref{ablation}, the results demonstrate the impact of these factors on the effectiveness of our framework. More ablations in Appendix D.

\begin{table}[t]
\centering
\vspace{-4mm}
\caption{Effect of Top-K and magnitudes on golf ball revival.}
\label{ablation}

\scalebox{0.85}{%
\begin{tabular}{lccc}
\toprule
\textbf{\shortstack{Top-K\\Retention}} &
\textbf{\shortstack{Neuron\\Sample}} &
\textbf{\shortstack{Neuron\\Average}} &
\textbf{\shortstack{NMS\\(Ours)}} \\
\midrule
Top 1.0 & 0.25 & 0.24 & 0.46 \\ 
Top 0.8 & 0.34 & 0.33 & 0.75  \\
Top 0.6 & 0.26 & 0.25 & 0.56 \\
Top 0.4 & 0.26 & 0.25 & 0.58  \\
Top 0.2 & 0.19 & 0.21 & 0.34 \\
\bottomrule
\end{tabular}
\vspace{-6mm}
}
\end{table}

\subsection{Limitations, Discussions, and Future Work}
\label{limits}
Although this work provides the first investigation into the security risks of pruning-based unlearning, several limitations remain and open new directions for future research. We focus only on single-concept unlearning, while future work can explore how multiple forgotten concepts interact and how revival occurs when several concepts are removed simultaneously. Our analysis assumes a white-box setting with full weight access; whether similar vulnerabilities exist under black-box conditions remains an open question. Moreover, our attack presumes the erased concept is known in advance. Future studies may investigate whether an attacker could infer the category of a forgotten concept without such prior knowledge, further revealing potential information leakage risks. We hope this work provides valuable insights and inspires future research on the security and reliability of machine unlearning.

\section{Conclusion}
This work reveals a previously overlooked security vulnerability in pruning-based unlearning for diffusion models. We demonstrate that pruning footprints can leak critical information, enabling concept revival even without data or retraining, and propose a simple yet effective defense to conceal such traces. Our findings highlight the need to reconsider the security assumptions of unlearning and inspire future research on safer and more reliable generative models.
\section*{Acknowledgment}
This material is partially based upon work supported by the National Science Foundation under Grant No. 2507128 and No. 2223768. 
Any opinions, findings, and conclusions or recommendations expressed in this material are those of the author(s) and do not necessarily reflect the views of the National Science Foundation.

{\small
\bibliographystyle{ieeenat_fullname}
\bibliography{main}
}

\clearpage
\setcounter{page}{1}
\maketitlesupplementary

\section*{A. Importance of Signs vs. Magnitudes in Concept Revival}
\label{signsvsmagnitudes}

Under the ideal assumption that we have access to the weights of the pretrained model. We conduct experiments to analyze the importance of pretrained magnitudes and pretrained signs in reviving visual concepts. Specifically, we design three revival strategies: pretrained magnitudes with signs sampled from the distribution of the remained weight neuron, pretrained signs with sampled magnitudes via the same distribution and seed as the former strategy, and pretrained signs with our neuron max scaling. The results are shown in Table \ref{ideal-object-reviving}. The experimental results reveal a clear trend. When pretrained magnitudes are paired with sampled signs, the revived concepts exhibit even worse recognition accuracy than the unlearned model. In contrast, using pretrained signs together with sampled magnitudes yields a substantially stronger revival effect, demonstrating that correct signs are far more critical than correct magnitudes in reviving erased visual concepts.


\begin{table}[H]
\centering
\vspace{-4mm}
\footnotesize
\renewcommand{\arraystretch}{1.0}
\caption{NMS Attack on Scissorhands and SalUn.}
\label{ideal-object-reviving}
\resizebox{\columnwidth}{!}{
\begin{tabular}{lccc||ccc}
\toprule
\multirow{2}{*}{Classes} &
\multirow{2}{*}{\shortstack{Scissorhands\\Unlearn $\downarrow$}} &
\multirow{2}{*}{\shortstack{Scissorhands\\NMS Attack $\uparrow$}} &
\multirow{2}{*}{\shortstack{Preserved Class\\Accuracy $\uparrow$}} &
\multirow{2}{*}{\shortstack{SalUn\\Unlearn $\downarrow$}} &
\multirow{2}{*}{\shortstack{SalUn\\NMS Attack $\uparrow$}} &
\multirow{2}{*}{\shortstack{Preserved Class\\Accuracy $\uparrow$}} \\
& & & & \\
\midrule
Church & 0.25 & 0.68 & 0.92 & 0.24 & 0.66 & 0.92 \\
Golf Ball & 0.23 & 0.76 & 0.93 & 0.24 & 0.80 & 0.91 \\
Gas Pump & 0.29 & 0.71 & 0.93 & 0.31 & 0.74 & 0.90 \\
Parachute & 0.20 & 0.62 & 0.91 & 0.21 & 0.65 & 0.92 \\
Chain Saw & 0.13 & 0.48 & 0.95 & 0.13 & 0.50 & 0.95 \\
French Horn & 0.13 & 0.34 & 0.90 & 0.14 & 0.44 & 0.91 \\
Mountain Bike & 0.11 & 0.67 & 0.93 & 0.13 & 0.71 & 0.90 \\
Starfish & 0.37 & 0.92 & 0.91 & 0.35 & 0.91 & 0.93 \\
Spider Web & 0.26 & 0.63 & 0.92 & 0.25 & 0.68 & 0.92 \\
School Bus & 0.15 & 0.79 & 0.93 & 0.18 & 0.83 & 0.91 \\
Racket & 0.21 & 0.68 & 0.91 & 0.20 & 0.66 & 0.91 \\
Candle & 0.17 & 0.66 & 0.92 & 0.14 & 0.63 & 0.92 \\
\midrule
Average & 0.21 & 0.66 & 0.92 & 0.21 & 0.68 & 0.92 \\
\bottomrule
\end{tabular}}
\end{table}

\section*{B. Why we do Top-K Sign Retention}

We observe that on the task of reviving erased concepts, the recovered weights across all visual concepts exhibit the same distribution pattern under our matrix completion method. Therefore, we select the recovered weights of \textit{Golf Ball} as the representative example for analysis. We divide recovered weights into five equal-magnitude partitions, denoted as $R_1$--$R_5$ from largest to smallest, and similarly divide the pretrained weights into five groups, $P_1$--$P_5$, where a larger index corresponds to smaller magnitudes. To quantify their alignment, we compute, for each pair $(R_i, P_i)$, the fraction of weight positions that overlap between the two groups, normalized by the number of $R_i$. 

The resulting alignment matrix is reported in Table~\ref{tab:alignment_matrix}.

\begin{table}[H]
\centering
\renewcommand{\arraystretch}{0.95}
\resizebox{0.8\columnwidth}{!}{%
\begin{tabular}{c|ccccc}
\hline
      & P1 & P2 & P3 & P4 & P5 \\ \hline
R1    &  \textbf{0.46} & 0.27 & 0.16 & 0.07 & 0.01 \\
R2    &  0.21 & \textbf{0.26} & 0.23 & 0.19 & 0.09 \\
R3    &  0.14 & 0.19 & \textbf{0.23} & 0.24 & 0.19 \\
R4    &  0.10 & 0.15 & 0.19 & \textbf{0.24} & 0.29 \\
R5    &  0.09 & 0.13 & 0.17 & 0.22 & \textbf{0.36} \\
\hline
\end{tabular}
}
\caption{Alignment matrix between recovered groups ($R_1$--$R_5$) and pretrained groups ($P_1$--$P_5$).}
\label{tab:alignment_matrix}
\end{table}

We observe that each recovered magnitude group $R_i$ exhibits a high positional overlap with its corresponding pretrained group $P_i$. \textbf{This indicates that our method effectively reconstructs the original magnitude structure of the pretrained model, placing large values back to positions that originally carried large weights.}

We further conduct a complementary analysis to examine how magnitude relates to sign correctness during recovery. Specifically, for each ratio $K \in \{0.2, 0.3, 0.5, 0.7, 0.9\}$, we select the top-$K$ recovered weights by magnitude as the \textbf{Top-K Magnitudes} group, and treat the remaining weights as the \textbf{Rest} group. By computing the sign accuracy within each group, we observe a clear trend: \textbf{weights with larger recovered magnitudes exhibit substantially higher sign correctness}. 

The quantitative results are presented in Table~\ref{tab:topk_rest}.

\begin{table}[H]
\centering
\renewcommand{\arraystretch}{1.15}
\begin{tabular}{c|c|c}
\hline
\multirow{2}{*}{K} 
    & \multicolumn{1}{c|}{Top-K Magnitudes'} 
    & \multicolumn{1}{c}{Rest (1-K)} \\
& \multicolumn{1}{c|}{Signs Accuracy} 
& \multicolumn{1}{c}{Signs Accuracy} \\ \hline
0.2 & 0.96 & 0.67 \\
0.3 & 0.94 & 0.64 \\
0.5 & 0.88 & 0.58 \\
0.7 & 0.81 & 0.54 \\
0.9 & 0.75 & 0.51 \\ \hline
\end{tabular}
\caption{Sign accuracy of recovered weights as a function of magnitude: Top-K groups consistently exhibit higher correctness than the Rest.}
\label{tab:topk_rest}
\end{table}

Combining these findings, matrix completion not only reconstructs the original magnitude structure of the pretrained model, but the recovered large-magnitude weights also exhibit significantly higher sign correctness. This makes \textbf{Top-K Sign Retention} a natural and necessary strategy for improving revival performance.

\section*{C. NSFW Content Revival}
For the task of nudity revival, we evaluate our method on I2P~\cite{schramowski2023safe}, MMA~\cite{yang2024mma} and Ring-A-Bell~\cite{tsai2023ring} datasets under three Top-K Sign Retention settings: $K = 0.4$, $0.6$, and $0.8$. The results are summarized in Table~\ref{tab:nudity_revival}.



\begin{table}[H]
\centering
\renewcommand{\arraystretch}{1.25} 
\begin{tabular}{cccc} 
\toprule
\multirow{2}{*}{\textbf{Method}}
& \multirow{2}{*}{\shortstack{\textbf{I2P} \\ (4703)}}
& \multirow{2}{*}{\shortstack{\textbf{MMA} \\ (1000)}}
& \multirow{2}{*}{\shortstack{\textbf{Ring-A-Bell} \\ (101)}} 
\\
& & & \\  
\midrule
\shortstack{Pretrained \\ SD-v1.5} & 461 & 785 & 101 \\
\shortstack{Concept \\ Prune}      & 74 & 57 & 22 \\
Top-0.4                            & 120 & 153 & 53 \\
Top-0.6                            & 78  & 173 & 46 \\
Top-0.8                            & 50 & 136 & 23 \\
\bottomrule
\end{tabular}
\caption{Nudity revival performance across I2P, MMA and Ring-A-Bell datasets under different Top-K settings.}
\label{tab:nudity_revival}
\end{table}

The three datasets I2P, MMA, and Ring-A-Bell contain 4703, 1000, and 101 prompts respectively. The number shown in the table corresponds to the number of images detected as containing NSFW content by the nudity detector~\cite{bedapudi2025nudenet}.

We observe that after applying Concept Prune~\cite{chavhan2024conceptprune}, the number of NSFW images generated by the model drops dramatically compared to the pretrained SD-v1.5.
However, after applying our revival framework, the number of images detected as NSFW increases substantially across all three datasets.

This clearly demonstrates that our method is highly effective at reviving the erased NSFW concepts. The visualization of revival in NSFW contents are shown in Fig~\ref{fig:nudity}

\section*{D. Ablation on Top-K Sign Retention and Magnitude Strategies}
We additionally perform ablation studies on the Parachute, Church, and Gas Pump revival tasks. The results are shown in Table~\ref{ablation_parachute}, Table~\ref{ablation_church} and Table~\ref{ablation_gas_pump} respectively.
\begin{table}[H]
\centering
\vspace{-4mm}
\caption{Effect of Top-K and magnitudes on parachute revival.}
\label{ablation_parachute}

\scalebox{0.85}{%
\begin{tabular}{cccc}
\toprule
\textbf{\shortstack{Top-K\\Retention}} &
\textbf{\shortstack{Neuron\\Sample}} &
\textbf{\shortstack{Neuron\\Average}} &
\textbf{\shortstack{NMS\\(Ours)}} \\
\midrule
Top 1.0 & 0.14 & 0.09 & 0.29 \\ 
Top 0.8 & 0.16 & 0.13 & 0.37  \\
Top 0.6 & 0.10 & 0.12 & 0.71 \\
Top 0.4 & 0.10 & 0.11 & 0.59  \\
Top 0.2 & 0.10 & 0.09 & 0.26 \\
\bottomrule
\end{tabular}
\vspace{-6mm}
}
\end{table}
\begin{table}[H]
\centering
\vspace{-4mm}
\caption{Effect of Top-K and magnitudes on church revival.}
\label{ablation_church}

\scalebox{0.85}{%
\begin{tabular}{cccc}
\toprule
\textbf{\shortstack{Top-K\\Retention}} &
\textbf{\shortstack{Neuron\\Sample}} &
\textbf{\shortstack{Neuron\\Average}} &
\textbf{\shortstack{NMS\\(Ours)}} \\
\midrule
Top 1.0 & 0.15 & 0.19 & 0.11 \\ 
Top 0.8 & 0.18 & 0.20 & 0.12  \\
Top 0.6 & 0.17 & 0.17 & 0.62 \\
Top 0.4 & 0.13 & 0.16 & 0.55  \\
Top 0.2 & 0.12 & 0.11 & 0.35 \\
\bottomrule
\end{tabular}
\vspace{-6mm}
}
\end{table}
\begin{table}[H]
\centering
\vspace{-4mm}
\caption{Effect of Top-K and magnitudes on gas pump revival.}
\label{ablation_gas_pump}

\scalebox{0.85}{%
\begin{tabular}{cccc}
\toprule
\textbf{\shortstack{Top-K\\Retention}} &
\textbf{\shortstack{Neuron\\Sample}} &
\textbf{\shortstack{Neuron\\Average}} &
\textbf{\shortstack{NMS\\(Ours)}} \\
\midrule
Top 1.0 & 0.12 & 0.13 & 0.01 \\ 
Top 0.8 & 0.16 & 0.15 & 0.05  \\
Top 0.6 & 0.15 & 0.13 & 0.30 \\
Top 0.4 & 0.10 & 0.11 & 0.37  \\
Top 0.2 & 0.10 & 0.10 & 0.24 \\
\bottomrule
\end{tabular}
\vspace{-6mm}
}
\end{table}

 Under a certain Top-K Sign Retention setting, the retained signs already capture the critical activation patterns associated with the erased visual concept. The results consistently show that our \textbf{Neuron Max Scaling} strategy further significantly amplifies such activation patterns by assigning magnitudes that reinforce retained signs. This leads to substantially improved revival performance compared to the other magnitude strategies. In addition, as the value of $K$ decreases, the revival performance exhibits a clear trend of rising first and then falling, indicating that shutting down too many activation channels deteriorates the model’s ability to reproduce the erased concept.

\section*{E. Attack diffusion unlearning methods that are based on targeting concept related weights}
To verify the generalization of effectiveness of sign-domain for concepts and our revival framework, we implement our revival framework on Scissorhands\cite{wu2024scissorhands} and SalUn\cite{fan2023salun} which both rely on gradients to identify concept-critical weights and then fine-tune those weights. Following their method, we prune those concept-critical weights and apply our NMS attack. As shown in Table \ref{ideal-object-reviving}, our NMS attack can effectively recover the unlearned accuracy from 21\% to 66\% and 68\%. And we conclude that once the positions of concept-related critical weights are aware, our method can successfully attack and recover the unlearned concept while preserving overall model utility.

\begin{table}[H]
\centering
\vspace{-4mm}
\footnotesize
\setlength{\tabcolsep}{15pt}  
\renewcommand{\arraystretch}{0.8}
\caption{Extend defense on all object erasure tasks}
\label{defense_extend_attack}
\resizebox{\columnwidth}{!}{
\begin{tabular}{lcccccc||cc}
\toprule
\multicolumn{1}{c}{\multirow{2}{*}{\shortstack[c]{\textbf{Gaussian}\\\textbf{Obfuscation}\\\textbf{\(\sigma_M\)}}}} &
\multicolumn{6}{c||}{\textbf{Extend Defense}} &
\multicolumn{2}{c}{\textbf{NMS Attack}} \\
\cmidrule(lr){2-7}\cmidrule(lr){8-9}
& \(10^{-6}\) & \(10^{-5}\) & \(10^{-4}\) & \(10^{-3}\) & \(10^{-2}\) & \(10^{-1}\)
& \(10^{-6}\) & \(10^{-2}\) \\
\midrule
Church & 0.08 & 0.09 & 0.09 & 0.08 & 0.06 & 0.00 & 0.06 & 0.08 \\
Golf Ball & 0.16 & 0.17 & 0.18 & 0.18 & 0.15 & 0.00 & 0.08 & 0.15 \\
Gas Pump & 0.07 & 0.07 & 0.09 & 0.05 & 0.03 & 0.00 & 0.01 & 0.07 \\
Parachute & 0.06 & 0.07 & 0.06 & 0.05 & 0.04 & 0.00 & 0.08 & 0.13 \\
Chain Saw & 0.00 & 0.02 & 0.04 & 0.04 & 0.05 & 0.03 & 0.01 & 0.01 \\
French Horn & 0.02 & 0.02 & 0.04 & 0.03 & 0.01 & 0.00 & 0.03 & 0.08 \\
Mountain Bike & 0.01 & 0.03 & 0.05 & 0.04 & 0.02 & 0.00 & 0.01 & 0.02 \\
Starfish & 0.32 & 0.30 & 0.27 & 0.25 & 0.20 & 0.14 & 0.04 & 0.31 \\
Spider Web & 0.22 & 0.25 & 0.27 & 0.20 & 0.16 & 0.05 & 0.03 & 0.18 \\
School Bus & 0.02 & 0.04 & 0.04 & 0.05 & 0.03 & 0.00 & 0.01 & 0.02 \\
Racket & 0.04 & 0.06 & 0.08 & 0.04 & 0.04 & 0.00 & 0.30 & 0.06 \\
Candle & 0.01 & 0.01 & 0.04 & 0.04 & 0.03 & 0.00 & 0.06 & 0.18 \\
\midrule
Average & 0.08 & 0.09 & 0.10 & 0.09 & 0.07 & 0.02 & 0.06 & 0.11 \\
\bottomrule
\end{tabular}}
\end{table}

\section*{F. Extensive evaluation on defense}
We extend our defense on all classes under different Gaussian Obfuscation. Moreover, we further apply the attack on the defended model. The results in the table \ref{defense_extend_attack} (Left) shows that our defense is effective on all classes under different Gaussian Obfuscation \(\sigma_M\). The results in the table \ref{defense_extend_attack} (Right) shows that our defense provides good resistance.






\begin{figure*}[t]
    \centering
    \includegraphics[width=0.9\textwidth]{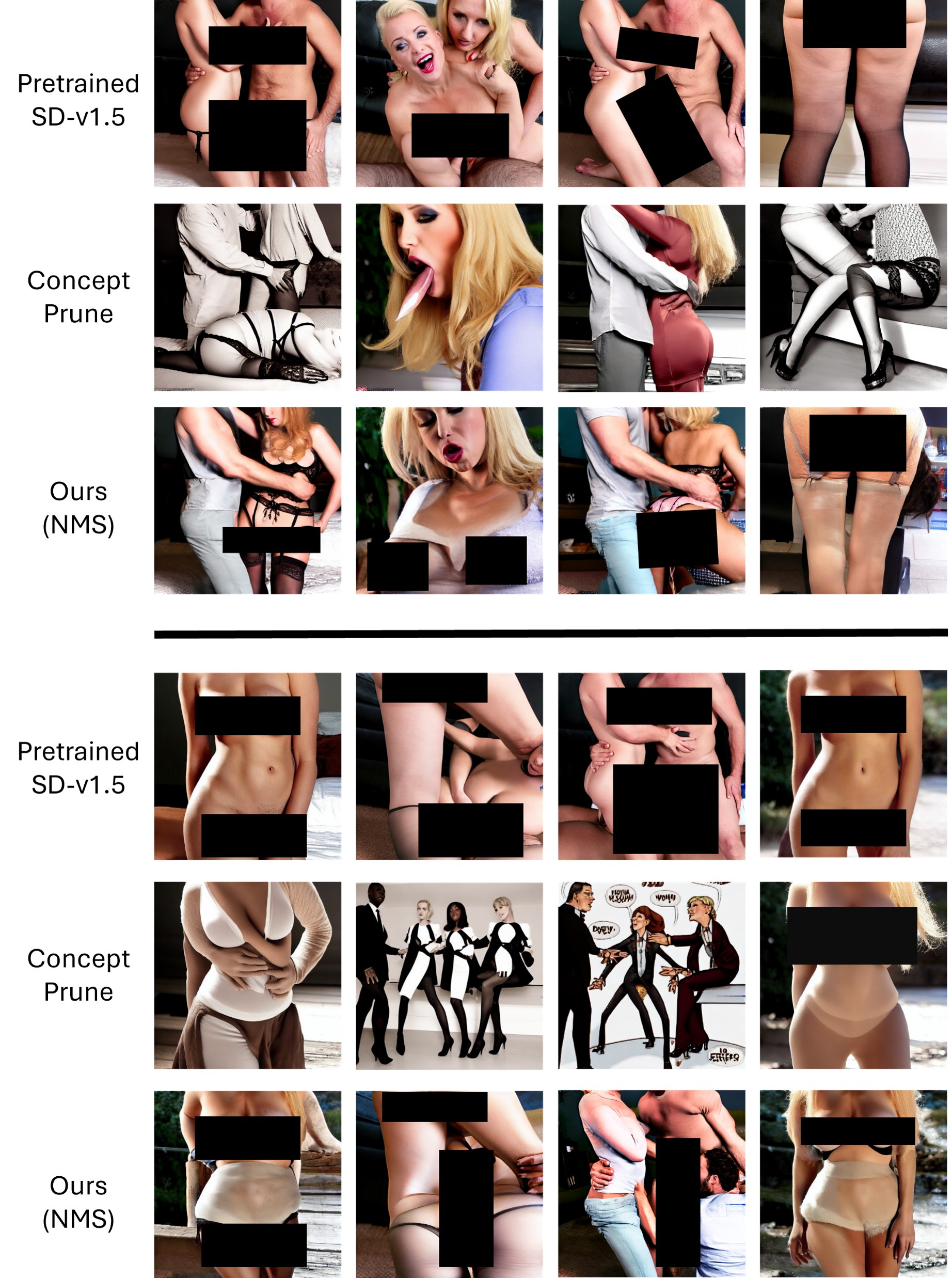}
    \caption{More visual results of revival on NSFW contents.}
    \label{fig:nudity}
\end{figure*}

\begin{figure*}
    \centering
    \includegraphics[width=1.0\textwidth]{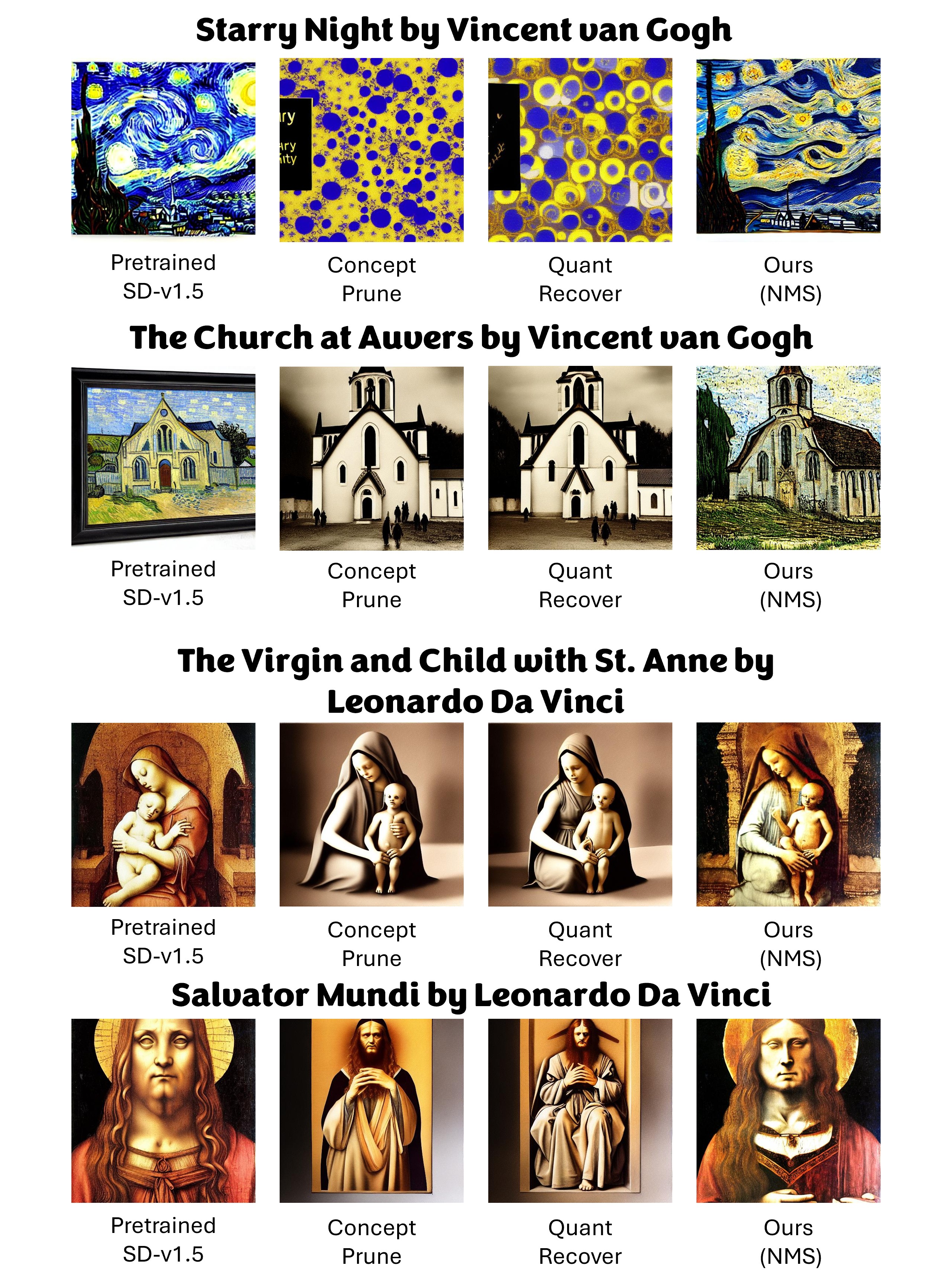}
    \caption{More visual results of revival on Van Gogh and Davinci.}
    \label{fig:nudity}
\end{figure*}

\begin{figure*}
    \centering
    \includegraphics[width=0.87\textwidth]{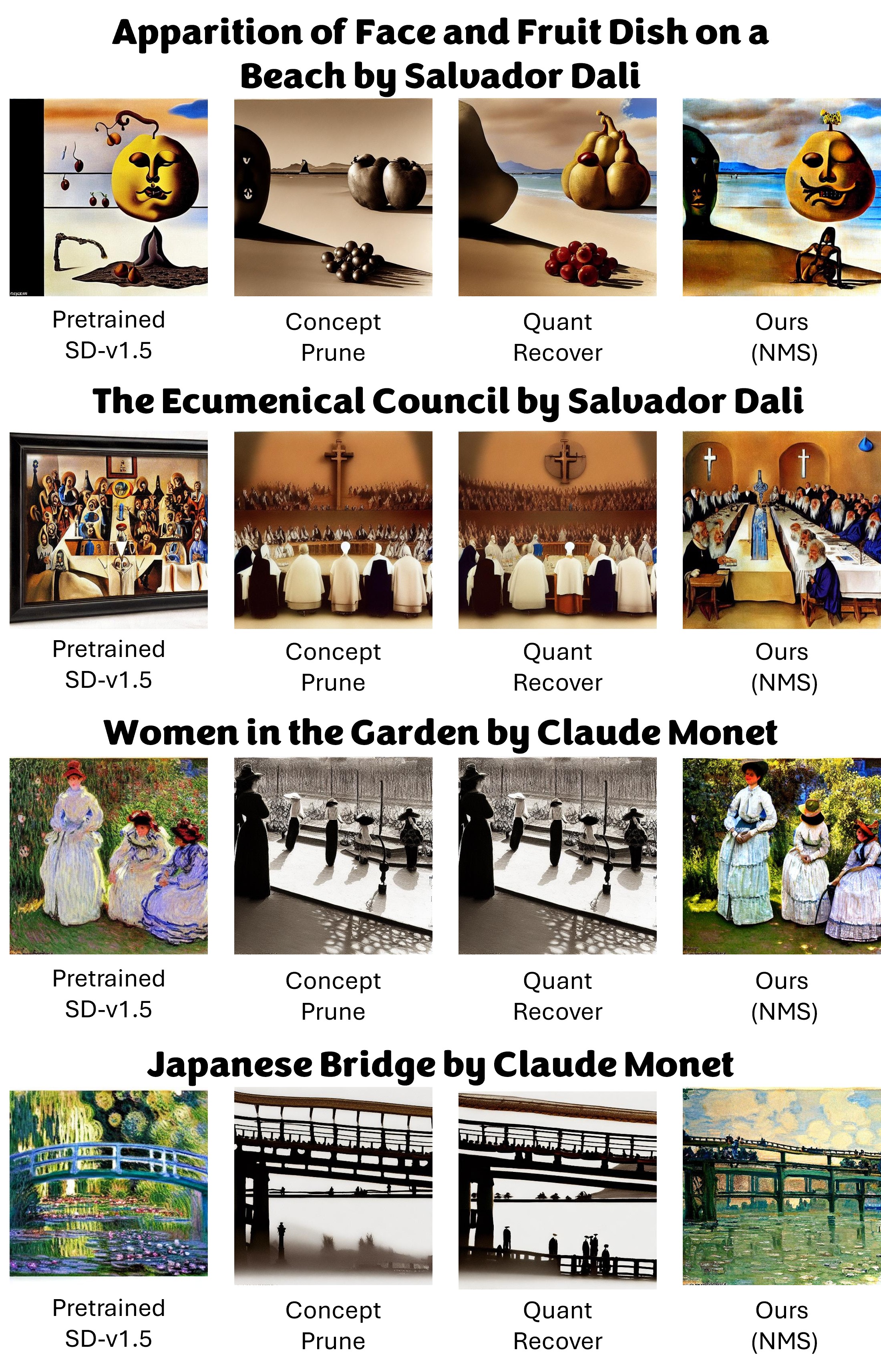}
    \caption{More visual results of revival on Dali and Monet.}
    \label{fig:nudity}
\end{figure*}

\begin{figure*}
    \centering
    \includegraphics[width=0.8\textwidth]{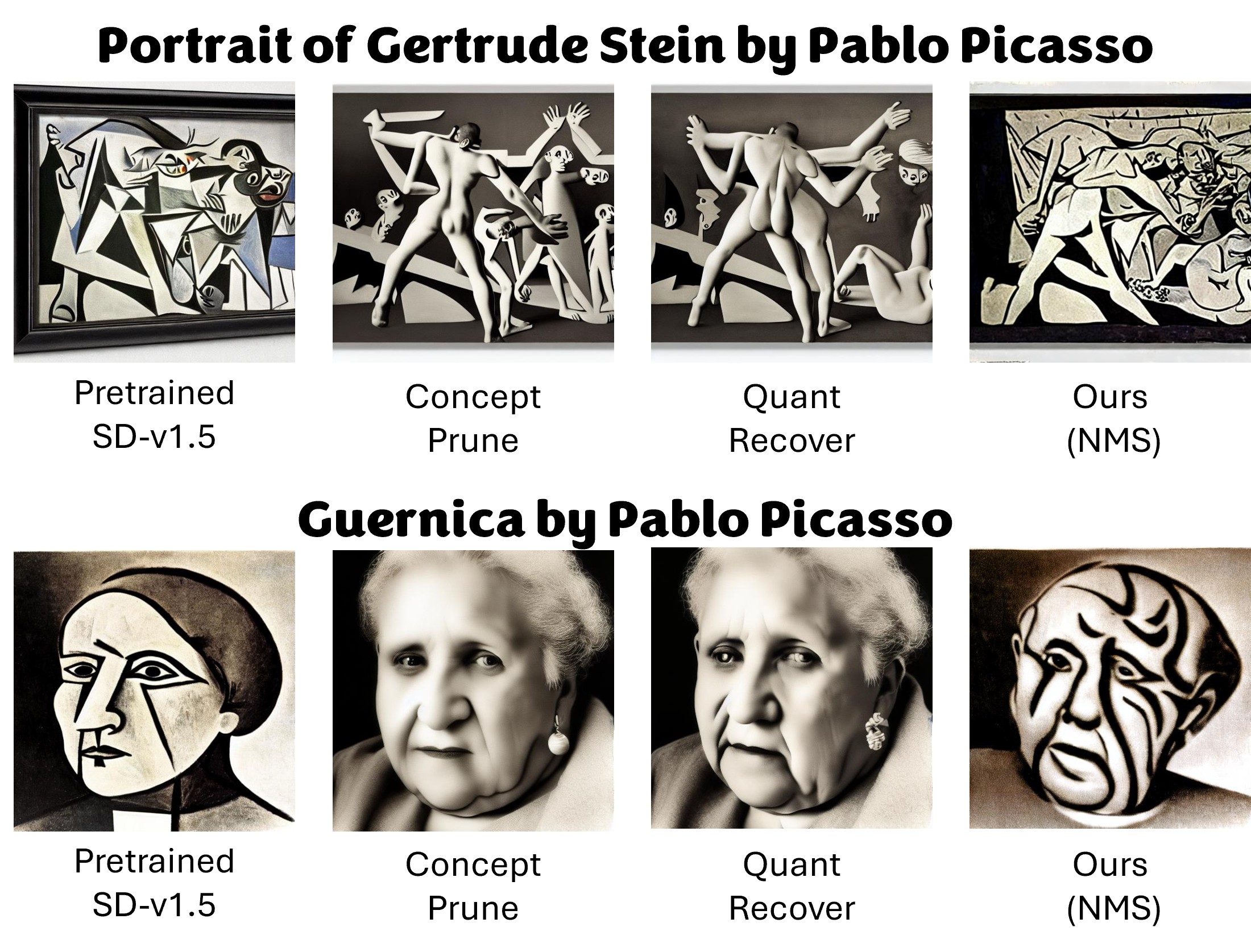}
    \caption{More visual results of revival on Picasso.}
    \label{fig:nudity}
\end{figure*}

\begin{figure*}[t]
    \centering
    \begin{minipage}{0.7\textwidth}
        \hspace{-9mm} 
        \includegraphics[width=\textwidth]{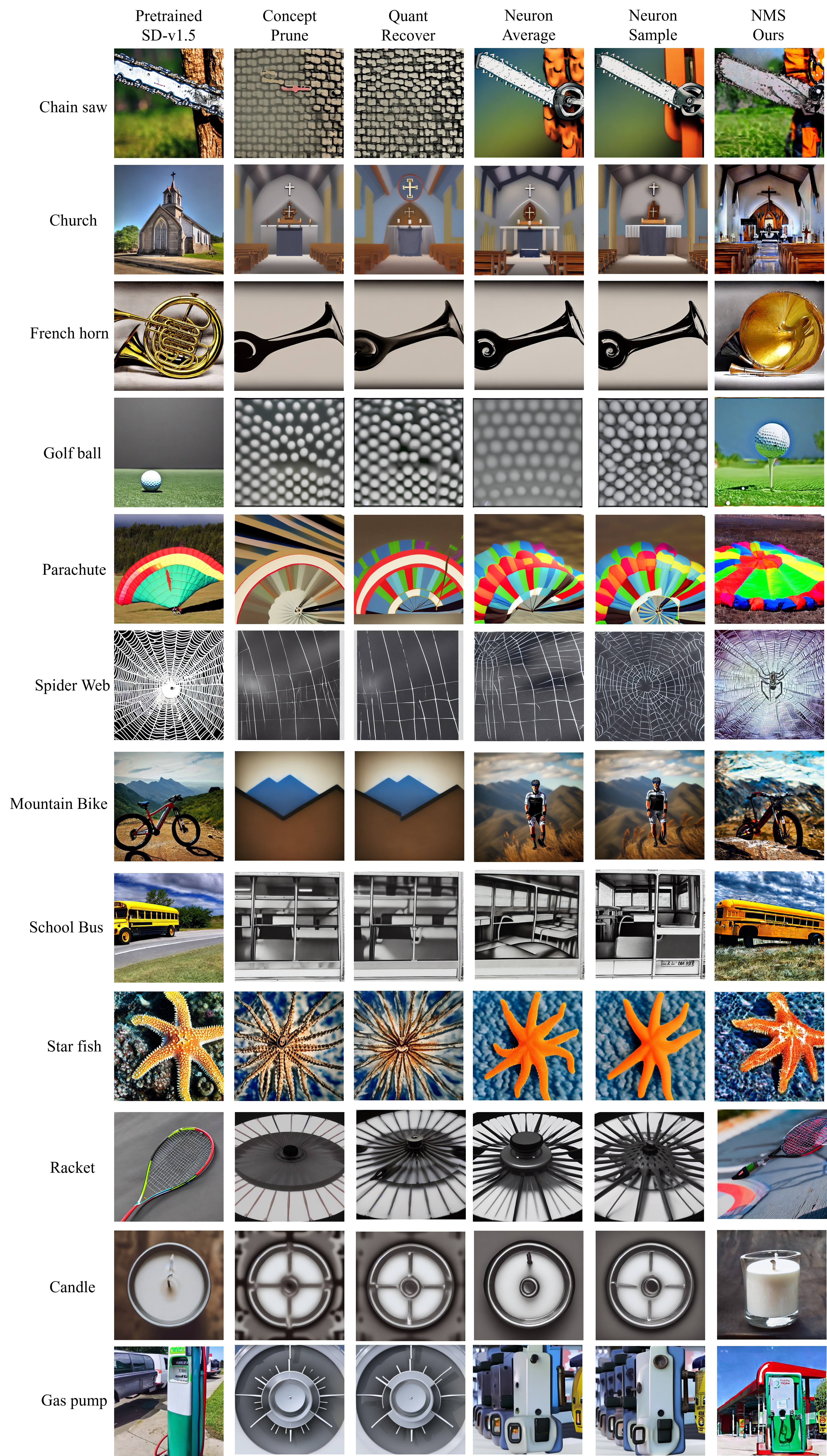}
    \end{minipage}
    \caption{More visual results of revival on erased objects.}
    \label{fig:nudity}
\end{figure*}

    







\end{document}